\documentclass[lettersize,journal]{IEEEtran}
\usepackage{cite}
\usepackage{amsmath,amssymb,amsfonts}
\usepackage{mathtools}
\usepackage{algorithm}
\usepackage{algorithmic}
\usepackage{graphicx}
\usepackage[caption=false,font=normalsize,labelfont=sf,textfont=sf]{subfig}
\usepackage{booktabs}
\usepackage{multirow}
\usepackage[table]{xcolor}
\usepackage{pifont}
\usepackage{textcomp}
\usepackage{stfloats}
\usepackage{url}
\usepackage[hidelinks]{hyperref}
\usepackage{tabularx}
\newcommand{\cmark}{\ding{51}}
\newcommand{\xmark}{\ding{55}}
\definecolor{tspgreen}{RGB}{226,240,217}
\definecolor{cvrpblue}{RGB}{221,243,250}

\def\BibTeX{{\rm B\kern-.05em{\sc i\kern-.025em b}\kern-.08em
    T\kern-.1667em\lower.7ex\hbox{E}\kern-.125emX}}
\begin{document}

\title{Rethinking Efficiency in Neural Combinatorial Optimization: Batched Preference Optimization with Mamba}
\author{Zhenxing Xu, Zeyuan Ma, Weidong Bao, Yan Zheng, Chongshuang Hu, Ji Wang and Zhiguang Cao
\thanks{Zhenxing Xu and Zeyuan Ma contributed equally to this work. Corresponding author: Weidong Bao.}
\thanks{Zhenxing Xu, Weidong Bao, Yan Zheng, Chongshuang Hu and Ji Wang are with the National Key Laboratory of Big Data and Decision, National University of Defense Technology (e-mail: wdbao@nudt.edu.cn).}
\thanks{Zeyuan Ma is with the School of Computer Science, South China University of Technology.}
\thanks{Zhiguang Cao is with the School of Computing and Information Systems, Singapore Management University.}}

\maketitle
\begin{abstract}
We study efficiency as a first-class objective in Neural Combinatorial Optimization (NCO) and present ECO, an efficient learning framework that combines batched preference optimization with a Mamba backbone. Instead of tightly interleaving every policy update with on-policy rollouts, ECO decouples trajectory generation from gradient updates through two stages: supervised warm-up on pre-computed solutions and iterative Direct Preference Optimization (DPO) on batched candidate sets generated by the current policy. We pair this learning pipeline with a mixed Mamba encoder-decoder that reduces encoder memory growth and recurrent decoding-state cost on long sequences while retaining the standard autoregressive candidate-scoring step. A local-search-guided bootstrapping strategy is further used during training to widen preference margins and stabilize iterative improvement. Importantly, local search is only used to construct stronger preference pairs during training and is never invoked at inference time. On TSP and CVRP, ECO is best or competitive across most compared neural baselines, with clear advantages in memory usage and throughput. We provide additional analysis on complexity decomposition, permutation sensitivity, evaluation fairness, and the contribution of each design component.

\end{abstract}

\section{Introduction}
\IEEEPARstart{C}{ombinatorial} Optimization (CO) is essential for routing, logistics, scheduling, and resource-allocation domains~\cite{tcyb-mra-vrp,tcyb-hcvrp,tcyb-csp,tcyb-moco}. Over the past decades, how to solve CO has been discussed and explored extensively, and representative legacy solvers such as Concorde~\cite{concorde2006}, Lin-Kernighan-Helsgaun algorithm~\cite{lkh}, and hybrid genetic search~\cite{HGS} more or less address CO's complexity. However, these meta-heuristics closely depend on expert-level knowledge to be crafted or even customized per case. As foretold by \emph{no-free-lunch} theorem~\cite{no-free-lunch}, such dependency may result in limited adaptability and scalability across diverse CO problems.

Learning-based approaches are continuously getting popular more recently for dealing with complex CO problems~\cite{2,tcyb-morl}, which are commonly termed as Neural Combinatorial Optimization (NCO). Along NCO's development, a crystal clear evolution path on the performance side could be observed from recent advances in this field: from pointer network~\cite{1} to Transformer architecture~\cite{transformer} and Generative Flow Network~\cite{flow-network}, from normal scale case~\cite{3} to massive scale generalization~\cite{boosting} and from simple reinforcement learning~\cite{4} to enhanced hybrid training~\cite{GFlowNet}. In contrast, comparatively limited attention has been paid to the efficiency side~\cite{efficient-nco-1,efficient-nco-2}. This motivates the core question of this paper: how can we improve NCO efficiency without giving up too much solution quality? This question matters because modern NCO training often requires millions of autoregressive rollouts, so poor convergence efficiency directly slows down research iteration, increases hardware cost, and limits practical deployment on larger instances.


From an efficiency perspective, the bottlenecks of existing NCO works predominantly come from two aspects: 1) \textbf{Learning Paradigm}: rollout collection and parameter updates are usually tightly coupled in an online loop. Such sample-then-update training under-utilizes GPUs and makes every optimization step depend on fresh autoregressive generation; 2) \textbf{Neural Network Architecture}: state-of-the-art NCO models predominantly rely on the Transformer architecture~\cite{transformer}. Although optimized kernels such as FlashAttention~\cite{flashattention} substantially improve constants and memory traffic, the underlying token-token interaction remains quadratic in sequence length, which becomes restrictive on long routing instances. At the same time, constructive routing still requires scoring feasible candidate nodes at each autoregressive step; therefore, architectural efficiency claims must distinguish sequence modeling, recurrent state updates, masking, and action scoring. This paper targets the former two bottlenecks while explicitly retaining candidate scoring in the constructive decoder.

To address this, we propose an efficiency-oriented NCO framework, termed \textbf{ECO}, that revisits both the learning pipeline and the backbone architecture. On the optimization side, ECO decouples trajectory generation from gradient updates through a two-stage batched preference learning workflow: a supervised warm-up stage on pre-computed trajectories, followed by iterative Direct Preference Optimization~(DPO)~\cite{10} on preference pairs refreshed between update rounds. We refer to this as a batched or offline-in-update paradigm rather than strict offline RL, because the dataset is periodically refreshed by the current policy. On the architecture side, we pair this training workflow with a tailored Mamba-based encoder-decoder~\cite{mamba} to reduce memory pressure and improve hardware utilization on long sequences. Without changing the autoregressive sampling protocol at test time, the resulting framework improves scalability and throughput via hardware-aware scan kernels and light-weight recurrent state updates. As illustrated in Figure~\ref{fig:intro}, ECO converges much faster than standard online baselines under the same hardware constraints while preserving competitive large-scale solution quality. Our contributions are three-fold:

\begin{figure}[t]  
\centering
\includegraphics[width=0.85\linewidth]{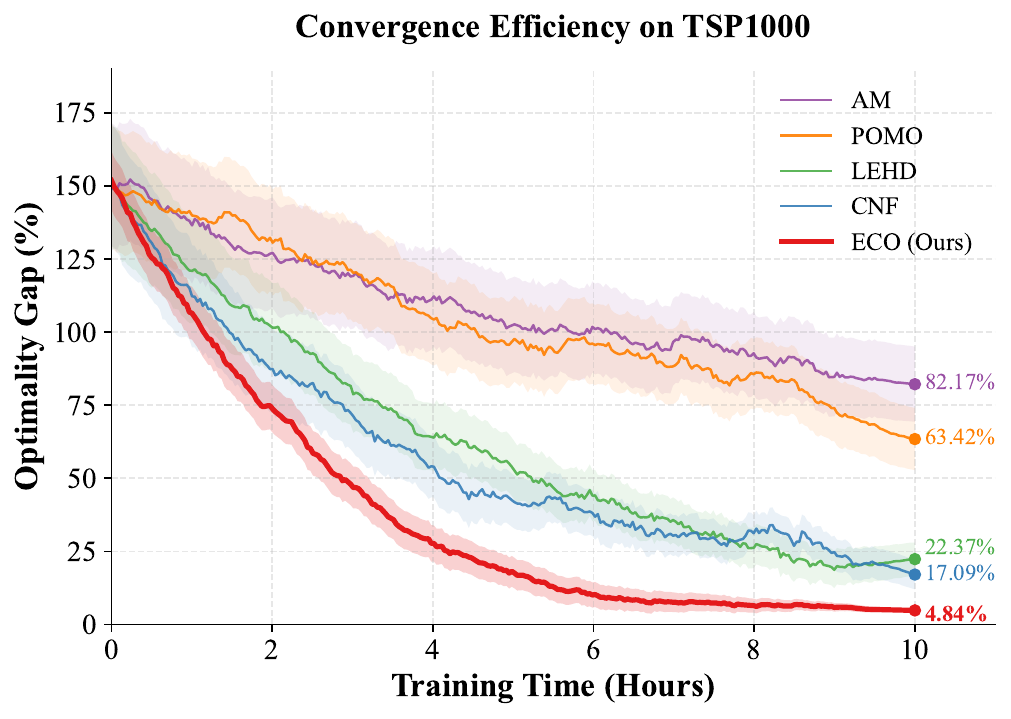}
\vspace{-5pt}
\caption{Convergence efficiency on TSP1000. The curves display the average optimality gap over 20 independent runs, with shaded regions indicating the standard deviation. All methods were trained on a single NVIDIA A800 GPU; wall-clock time includes rollout or candidate generation during training but excludes the one-time generation of shared SFT labels. Batch sizes were tuned within the same memory budget for each method.}
\label{fig:intro}
\vspace{-5pt}
\end{figure}

\begin{itemize}

    \item[\ding{182}]  To our knowledge, we introduce the first Mamba-based NCO solver and provide an explicit complexity decomposition showing that Mamba reduces encoder memory scaling and recurrent decoding-state cost, while naive autoregressive candidate scoring remains quadratic in the number of nodes.

    \item[\ding{183}]  We introduce an efficiency-oriented NCO framework that combines a tailored Mamba architecture, a two-stage batched preference optimization pipeline, and a local-search-guided bootstrapping strategy to stabilize training and amplify useful preference margins.

    \item[\ding{184}]  We conduct case studies on TSP and CVRP, where benchmarking, memory, throughput analysis, permutation tests, and ablations show that ECO delivers best-or-competitive neural performance together with substantial efficiency gains, especially on large instances.
    \vspace{-7pt}

\end{itemize}

\section{Related Works}

\subsection{Neural Combinatorial Optimization and Scalability Bottlenecks}

NCO has evolved from Pointer Networks~\cite{1} and RL-based solvers~\cite{2} to attention-based models such as AM~\cite{3} and POMO~\cite{4}. Prior studies have also explored deep-reinforcement-learning solvers for VRP variants, covering salesman problems, and multiobjective CO~\cite{tcyb-mra-vrp,tcyb-hcvrp,tcyb-csp,tcyb-morl,tcyb-moco}. Recent scalable variants, including LEHD~\cite{5}, Boosting~\cite{boosting}, GLOP~\cite{glop}, BQ~\cite{bq}, heatmap-guided solvers such as Att-GCN+MCTS~\cite{attgcn_mcts}, diffusion solvers such as DIFUSCO~\cite{difusco}, and neural improvement methods such as NeuOpt~\cite{neuopt}, further improve large-scale routing from complementary directions. These methods differ substantially in inference protocol: some are constructive policies, while others rely on heatmap decoding, partitioning, diffusion refinement, augmentation, or iterative local improvement. We therefore keep the main quantitative comparison focused on constructive neural baselines with compatible end-to-end inference and discuss the broader scalable-routing literature to clarify the scope of the comparison. Long-sequence NCO is still bottlenecked by expensive online rollouts and the memory/latency cost of attention. Even with optimized kernels such as FlashAttention~\cite{flashattention}, constructive attention models retain quadratic token interactions. This motivates more hardware-efficient architectures and training pipelines.
\vspace{-10pt}
\subsection{Sequence Modeling via Mamba}\label{sec:2.2}
Structured state-space models (SSMs) have emerged as an efficient alternative to attention for long-sequence modeling~\cite{6,ssm-gu,h3,linear-attn,retnet}. For an input sequence $x \in \mathbb{R}^{L \times D}$, an SSM updates a hidden state $h(t)$ and output $y(t)$ as:
\begin{equation}\label{eq:s4}
    h(t) = \overline{A} h(t-1) + \overline{B} x(t),\quad y(t) = C h(t).
\end{equation}
In the time-invariant S4 form, training can be parallelized by global convolution:
\begin{equation}\label{eq:s4-kernal}
    \overline{K} = (C\overline{B},C\overline{AB},C\overline{A}^2\overline{B},...),\quad Y = X\cdot \overline{K}.
\end{equation}
This gives S4 efficient parallel training and linear-recurrence inference. Mamba~\cite{mamba,mamba2} extends SSMs with input-dependent parameters, improving selective sequence modeling while preserving efficient scan-based computation. These properties make it attractive for long-sequence NCO, where memory and throughput are critical.

\begin{figure*}[!ht]  
\centering
\includegraphics[width=0.98\linewidth]{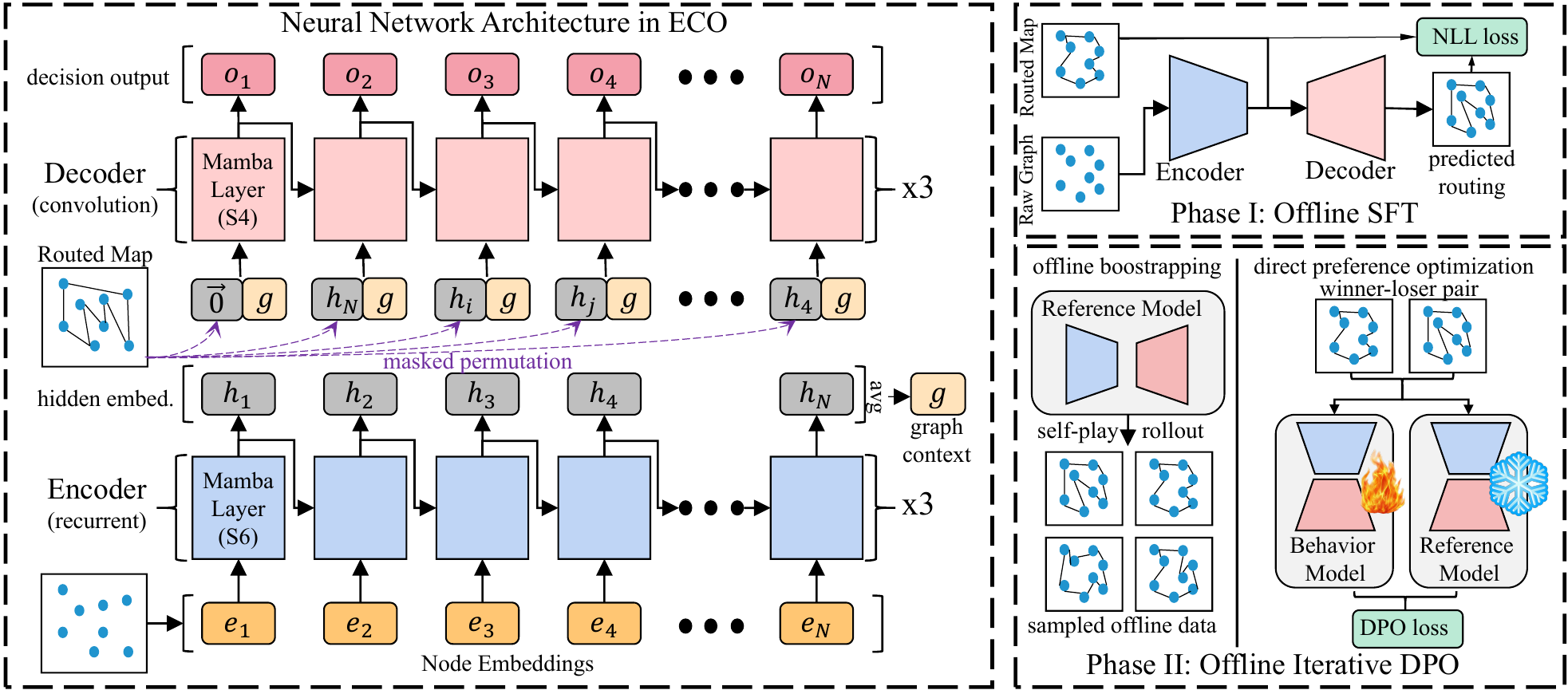}

\caption{The core designs and workflow of our ECO. \textbf{Left}: The Mamba-based Encoder-Decoder architecture proposed to facilitate offline NCO training. \textbf{Right}: The two-phase offline learning with a SFT warm-up as quick knowledge adaption and an iterative preference learning to further enhance performance.}
\label{fig:method}
\vspace{-15pt}
\end{figure*}
\vspace{-5pt}
\subsection{Preference Learning}


DPO~\cite{10} optimizes a policy from pairwise preferences without explicit reward modeling, and related works such as ReST~\cite{11} and Self-Rewarding LM~\cite{12} show the value of iterative self-improvement. In NCO, the reward is deterministic and cheap to compute, but DPO is still attractive because it converts absolute costs into a pairwise supervised objective, decouples data generation from optimization, and reduces the variance of online policy gradients. Concurrent works have already explored DPO for NCO~\cite{dpo-nco-1,dpo-nco-2}; our focus is its integration with a Mamba backbone and LS-guided bootstrapping for efficient training.

\section{Method}

The core idea of ECO is illustrated in Figure~\ref{fig:method}. ECO follows a batched preference-optimization paradigm for efficient NCO training: policy updates are performed on buffered trajectory pairs, while new rollouts are generated only between update rounds. On top of this pipeline, we introduce a mixed Mamba architecture that targets long-sequence modeling with lower memory pressure. We further design a two-stage training workflow to improve robustness in this setting. In the subsequent sections, we detail each part of ECO step by step.

\vspace{-5pt}
\subsection{Problem Formulation}
We consider a general class of combinatorial optimization problems that can be formulated as finding an optimal permutation or sequence of nodes on a graph. Let $G = (V, E)$ be a graph where $V = \{v_1, \dots, v_N\}$ represents a set of $N$ nodes (e.g., cities in TSP, customers in CVRP) characterized by feature vectors $X = \{x_1, \dots, x_N\}$ (e.g., coordinates, demands). The goal is to generate a sequence $\pi = (\pi_1, \dots, \pi_T)$ of indices from $\{1, \dots, N\}$ that satisfies specific feasibility constraints $\Omega$ (e.g., visiting every node exactly once, capacity limits) while minimizing a cost objective $\mathcal{C}(\pi | G)$. The problem is formally defined as:

\begin{equation}
   \pi^* = \arg\min_{\pi \in \Omega} \mathcal{C}(\pi | G).
\end{equation}
We model this as a probabilistic sequence generation task using a parameterized policy $p_\theta(\pi | X)$. By decomposing the joint probability via the chain rule, the probability of a solution $\pi$ is:
\begin{equation}
    p_\theta(\pi | X) = \prod_{t=1}^T p_\theta(\pi_t | \pi_{<t}, X),
\end{equation}
where $\pi_{<t}$ denotes the partial solution constructed up to step $t-1$. Our objective is to optimize parameters $\theta$ such that the expected cost $\mathbb{E}_{\pi \sim p_\theta}[\mathcal{C}(\pi|X)]$ is minimized.
For CVRP, we serialize the multi-route solution as a single sequence by introducing a depot token that may appear multiple times. Each customer node contains coordinates and demand, while the dynamic state at step $t$ is summarized by a visited mask and the remaining vehicle capacity $q_t$. A customer $j$ is feasible only if it is unvisited and its demand satisfies $d_j \le q_t$; selecting the depot resets the remaining capacity to the vehicle capacity $Q$, and consecutive depot-to-depot transitions are masked out until all customers are served. This serialization allows TSP and CVRP to share a unified autoregressive formulation while preserving the task-specific feasibility constraints in $\Omega$.
\vspace{-5pt}
\subsection{Batched Preference Optimization Paradigm}\label{sec:3.3}

Traditional RL methods for CO, such as REINFORCE with a greedy rollout baseline, often suffer from high variance and training instability. Inspired by recent advances in preference optimization, we propose a two-stage strategy: a warm-up phase using supervised learning and a continuous policy improvement phase using iterative DPO~\cite{10}. The resulting pipeline is not strict offline RL because the preference buffer is periodically refreshed by the current policy; however, each optimization phase itself operates on pre-generated batches, which is the source of the efficiency gain we target.

\subsubsection{Phase I: Supervised Fine-Tuning (SFT)}

This phase gives the policy a basic understanding of routing constraints and local optimality through behavior cloning on high-quality solutions.

\textbf{Data Generation:} We generate a dataset $\mathcal{D}_{\text{SFT}}$ using the LKH-3 solver (or nearest neighbor heuristics for rapid initialization).

\textbf{Objective:} We minimize the Negative Log-Likelihood (NLL) of the expert trajectories $\pi^*$:

\begin{equation}
    \mathcal{J}_{\text{SFT}}(\theta) = - \mathbb{E}_{(X, \pi^*) \sim \mathcal{D}_{\text{SFT}}} \left[ \sum_{t=1}^T \log p_\theta(\pi_t^* | \pi_{<t}^*, X) \right]
\end{equation}

\subsubsection{Phase II: Iterative Direct Preference Optimization (DPO)}

The core of our training is the Iterative DPO algorithm. Unlike PPO, DPO optimizes the policy directly from preference data without training a separate value function (Critic), significantly reducing memory overhead and training complexity.

\textbf{Preference Pair Construction:} In each iteration $k$, we generate a dataset of preference pairs. For a given instance $X$, we sample $K$ solutions $\{y_1, \dots, y_K\}$ using the current policy $\pi_{\theta}$. To construct a pair $(y_w, y_l)$:
\textbf{Winner ($y_w$):} The trajectory with the shorter tour length. \textbf{Loser ($y_l$):} The trajectory with the longer tour length.

\textbf{DPO Objective:} We optimize the policy $\pi_\theta$ to maximize the margin between the likelihood of the winner and the loser, constrained by a reference model $\pi_{\text{ref}}$ (the policy from the previous iteration). The loss function is:
\vspace{-5pt}
\begin{equation}
\begin{split}
\mathcal{L}_{\text{DPO}}(\theta; \pi_{\text{ref}}) = 
- \mathbb{E}_{(X, y_w, y_l)} \bigg[ 
& \log \sigma \bigg( \beta \log \frac{\pi_\theta(y_w|X)}{\pi_{\text{ref}}(y_w|X)} \\
& - \beta \log \frac{\pi_\theta(y_l|X)}{\pi_{\text{ref}}(y_l|X)} \bigg) \bigg]
\end{split}
\end{equation}

where $\sigma$ is the sigmoid function and $\beta$ is a temperature parameter controlling the deviation from the reference model.

\textbf{Iterative Refinement:} Instead of a static reference model, we update $\pi_{\text{ref}} \leftarrow \pi_{\theta}$ every $T$ steps. This schedule keeps the KL anchor close to the current policy and reduces the mismatch between the optimization target and the sampled data. We do not claim a formal global improvement guarantee; rather, the design acts as a practical continuation strategy that empirically stabilizes training in our setting.

\subsubsection{Enhancing Preference Contrast via Local Search}

In the standard DPO framework, preference pairs are constructed solely from model-sampled trajectories. As training progresses, the policy tends to generate solutions with similar quality, resulting in narrow reward margins ($|L(y_w) - L(y_l)| \to 0$). This leads to vanishing gradients and hinders the model's ability to distinguish subtle structural improvements.

To address this, we introduce an LS-Augmented Preference Construction mechanism. This strategy integrates a lightweight local search operator $\mathcal{T}_{\text{LS}}$ (e.g., 2-opt or 3-opt) into the data generation process to amplify the quality contrast and distill local optimization capabilities into the neural policy.

Mechanism:
For a subset of instances in each iteration, given a model-generated solution $y_{\text{raw}} \sim \pi_\theta(\cdot|X)$, we apply the local search operator to obtain an improved version: $y_{\text{refined}} = \mathcal{T}_{\text{LS}}(y_{\text{raw}})$. We then construct the preference pair $(y_w, y_l)$ strictly as:
\vspace{-5pt}
\begin{equation}
    y_w = y_{\text{refined}}, \quad y_l = y_{\text{raw}}
\end{equation}

Since $\mathcal{T}_{\text{LS}}$ accepts only improving or equal-cost moves, it guarantees $L(y_{\text{refined}}) \le L(y_{\text{raw}})$ and provides a cleaner supervision signal.

\textbf{Margin Amplification:} By manually widening the performance gap between the winner and the loser, we increase the magnitude of the implicit reward difference $\hat{r}_\theta(y_w) - \hat{r}_\theta(y_l)$. This results in sharper gradients, accelerating convergence in plateau regions.

\textbf{Distilling Local Improvement Patterns into the Policy:} This process can be viewed as a distillation-style signal. By consistently preferring the LS-refined solution over the raw output, the model is encouraged to absorb recurring local improvement patterns (e.g., removing crossings or tightening short sub-tours). Importantly, in our framework local search is \emph{only} used during training for preference construction and is never applied during inference. Therefore, any performance gain of the full ECO model over the `w/o LS Boot.' variant in Table~\ref{tab:main_results} must come from improved network parameters rather than search-time post-processing. We still avoid claiming exact internalization of every LS operation; our claim is more modest: LS supplies a stronger training signal that helps the policy learn better constructive behaviors.

\subsection{Mixed Mamba-based Architecture}
Transformer-based NCO models (e.g., AM~\cite{3} and POMO~\cite{4}) remain challenging on long routing instances even when aided by software- and hardware-level optimizations such as KV cache~\cite{kv-cache} and FlashAttention~\cite{flashattention}. To address this, we propose a tailored neural network architecture (Figure~\ref{fig:method}) that uses 1) a recurrent-mode Mamba encoder to reduce memory usage while extracting long-range correlations, and 2) an S4-style decoder that supports parallel teacher-forced state computation during training and efficient recurrent updates at inference. This design enables the architecture and training pipeline to reinforce each other: the Mamba backbone improves scalability on long sequences, while the batched preference-learning setup fully exploits its parallelism and memory efficiency.

\subsubsection{Recurrent Mamba Encoder}

The encoder transforms the raw node set into contextual node embeddings. Unlike the standard Transformer encoder which computes an $N \times N$ attention map, our encoder utilizes stacked Mamba layers\footnote{https://github.com/state-spaces/mamba}. Since Mamba is a sequence model, the node order is an architectural choice rather than a harmless implementation detail. We therefore canonicalize the input before the Mamba scan. For TSP, customer nodes are sorted by a Morton (Z-order) code computed from their normalized 2-D coordinates; for CVRP, the depot is placed at position 0 and customers are sorted by the same Morton order. The policy always outputs the original node indices, so this ordering only affects the internal sequence representation. We choose Morton order because it is deterministic, cheap to compute, and preserves spatial locality. Sec.~\ref{sec:perm_sensitivity} reports a permutation-sensitivity study and ablations over alternative orderings.

\textbf{Input Projection:} Let $\rho$ denote the canonical ordering map. The ordered node feature $x_{\rho(i)}$ is projected into a $D$-dimensional latent space via a linear transformation:
\begin{equation}
    e_i = W_{\text{in}} x_{\rho(i)} + b_{\text{in}} .
\end{equation}
\textbf{Mamba Layers:} We use Mamba layers with S6 setting~(see the end of Sec.~\ref{sec:2.2}). This input-dependent setting helps the encoder to extract spatial dependence across a long sequence of nodes in a large scale CO problems. We denote the $L=3$ layers' parameters as $M_{enc}^{(1)}$, $M_{enc}^{(2)}$ and $M_{enc}^{(3)}$. Then following the per-layer computation in Eq.~\ref{eq:s4}, we recurrently compute hidden embeddings of all nodes:
\begin{equation}
    H = M_{enc}^{(3)} \circ M_{enc}^{(2)} \circ M_{enc}^{(1)}(E),
\end{equation}
where $E$ is the ordered embedding sequence and $H=\{h_i\}_{i=1}^{N}$ are the obtained hidden embeddings for all nodes in the CO problems. We additionally compute a global graph embedding $g = \operatorname{mean}(H)$ as context information used for the decoding phase. Since we adopt Mamba with the S6 setting, training uses the hardware-aware parallel scan and recurrent inference uses a fixed-size state.



\subsubsection{S4 Decoder for Teacher-Forced Training and Recurrent Inference}

\textbf{Training-time Input Preparation.} During SFT and DPO updates, the decoder is paired with routed trajectories from the current batch. Given the hidden embeddings $H=\{h_i\}_{i=1}^{N}$, we gather the embedding sequence along the teacher route $\pi=(\pi_1,\ldots,\pi_T)$. For TSP, $T=N$. For CVRP, $T=N+R$, where $R$ is the number of depot visits in the serialized multi-route solution, and hence $T$ can exceed $N$. We prepare the decoder input as:
\begin{equation}
    z_t = h_{\pi_{t-1}} + g + W_d d_t, \quad t=1,\ldots,T,
\label{eq:decoder_input}
\end{equation}
where $d_t$ denotes task-specific dynamic features, e.g., remaining capacity and a depot indicator for CVRP. We use a learned start token for $t=1$.

\textbf{Teacher-forced State Computation.} We use Mamba layers with S4 setting~(see Sec.~\ref{sec:2.2} and Eq.~\ref{eq:s4-kernal}) to compute decoder states in parallel during training. We denote the $L=3$ layers' parameters as $M_{dec}^{(1)}$, $M_{dec}^{(2)}$ and $M_{dec}^{(3)}$ and obtain decision output $O:\{o_i\}_{i=1}^{N}$ by:
\begin{equation}
    S = M_{dec}^{(3)} \circ M_{dec}^{(2)} \circ M_{dec}^{(1)}(Z),
\label{eq:decoder_states}
\end{equation}
where $S=\{s_t\}_{t=1}^{T}$ and $Z=\{z_t\}_{t=1}^{T}$. The action logits are then computed against the encoder-side candidate embeddings:
\begin{equation}
    \ell_{t,j}
    =
    \frac{(W_q s_t)^\top (W_k h_j)}{\sqrt{D}}
    + b_j,
    \quad j\in V_0,
\label{eq:action_logit}
\end{equation}
where $V_0=V$ for TSP and $V_0=V\cup\{0\}$ for CVRP with depot token $0$. This is a pointer-style compatibility score implemented as a bilinear dot product; it is not a self-attention layer because the candidate embeddings are not mutually attended during decoding. The probability of the next action is
\begin{equation}
    p_\theta(\pi_t=j|\pi_{<t},X)
    =
    \frac{\exp(\ell_{t,j})\mathbb{I}[j\in\mathcal{A}_t]}
    {\sum_{k\in V_0}\exp(\ell_{t,k})\mathbb{I}[k\in\mathcal{A}_t]},
\label{eq:masked_policy}
\end{equation}
where $\mathcal{A}_t$ is the feasible candidate set after applying the task mask.

\textbf{Autoregressive Inference.} At test time, no routed map is available beforehand. We therefore keep the standard autoregressive sampling paradigm. The decoder runs recurrently as $s_t=\operatorname{SSM}_{step}(s_{t-1},z_t)$, updates a constant-size hidden state from the previously selected node and dynamic features, and scores the next feasible action by Eq.~\eqref{eq:action_logit}. Hence, the benefit of the decoder at inference is lower memory pressure and cheaper recurrent state updates, not one-shot parallel generation of the entire route and not elimination of candidate scoring.

\textbf{Task-specific Masking.} For TSP, the feasibility mask simply forbids revisiting already selected cities. For CVRP, we additionally concatenate normalized demand to the node feature, include a dedicated depot token, and enforce the remaining-capacity constraint in the mask. A customer is available only when it is unvisited and its demand does not exceed the current remaining capacity. The depot token is never permanently marked as visited; selecting it closes the current route segment, resets the remaining capacity to $Q$, and is masked immediately after another depot unless all customers have been served.






\subsubsection{Algorithm Complexity}
\label{sec:complexity}
We decompose the complexity because a constructive solver contains several distinct operations. Let $D$ be the embedding dimension, $T$ be the decoded sequence length, and $|\mathcal{A}_t|$ be the number of feasible actions at step $t$. The Mamba encoder uses linear-memory sequence modeling, with $O(ND)$ activation memory up to constant state factors, whereas a standard full-attention encoder stores an $O(N^2)$ attention map. The recurrent decoder state update costs $O(TD)$ because the hidden state has fixed size. However, the action scorer in Eq.~\eqref{eq:action_logit} still compares the current decoder state with feasible candidate embeddings. With naive dense scoring, this costs
\begin{equation}
    O\!\left(D\sum_{t=1}^{T} |\mathcal{A}_t|\right),
\label{eq:scoring_complexity}
\end{equation}
which is $O(DN^2)$ for TSP because $T=N$ and $|\mathcal{A}_t|=N-t+1$. For CVRP, $T=N+R$ and the scoring cost is $O(DTN)$ in the worst case, where repeated depot decisions make $T$ larger than $N$. Mask and feasibility updates are $O(N)$ per step with a dense mask, or lower in practice with vectorized bit/boolean operations, but they do not change the worst-case dense-scoring order.

Therefore, ECO should not be interpreted as making complete constructive inference linear in $N$. Its architectural gain is more specific: it reduces encoder memory scaling and the recurrent decoding-state cost, improves GPU utilization through scan kernels and batched preference updates, and avoids decoder self-attention/KV-cache growth. The end-to-end naive autoregressive decoder remains quadratic because pointer-style action scoring over candidates is still required. Sec.~\ref{sec:4.2.2} empirically evaluates the resulting memory and throughput advantages under both fixed-batch and max-utilization protocols.

\section{Experiment}

\begin{table*}[t]
\centering
\caption{
\textbf{Performance comparison on TSP and CVRP instances of varying sizes.} 
We report the average objective value (Obj.), the average optimality gap (Gap) relative to the ground truth, and the total wall-clock evaluation time (Time) over the same 1,000-instance test set. Time follows the max-utilization batched protocol recommended by each implementation; fixed-batch timing and protocol details are reported in Tables~\ref{tab:fixed_batch_runtime} and~\ref{tab:eval_protocol}.
\textbf{Concorde}, \textbf{LKH-3} and \textbf{HGS} are used as the ground truth solvers for TSP and CVRP, respectively (indicated by 0.00\% Gap). 
`ECO' denotes the full model trained with LS-augmented preference construction, while `ECO w/o LS Boot.' removes this training-time signal. No local search is used during inference for either row. For learning-based methods, the best results are highlighted in \textbf{bold}, and the second-best results are \underline{underlined}.
}
\label{tab:main_results}
\resizebox{\textwidth}{!}{
\begin{tabular}{l|ccc|ccc|ccc|ccc}

\toprule
\multicolumn{1}{c|}{\multirow{2}{*}{Method}} & \multicolumn{3}{c|}{\cellcolor{tspgreen}TSP200} & \multicolumn{3}{c|}{\cellcolor{tspgreen}TSP500} & \multicolumn{3}{c|}{\cellcolor{tspgreen}TSP1000} & \multicolumn{3}{c}{\cellcolor{tspgreen}TSP5000} \\
 & \cellcolor{tspgreen}Obj.$\downarrow$ & \cellcolor{tspgreen}Gap$\downarrow$ & \cellcolor{tspgreen}Time$\downarrow$ & \cellcolor{tspgreen}Obj.$\downarrow$ & \cellcolor{tspgreen}Gap$\downarrow$ & \cellcolor{tspgreen}Time$\downarrow$ & \cellcolor{tspgreen}Obj.$\downarrow$ & \cellcolor{tspgreen}Gap$\downarrow$ & \cellcolor{tspgreen}Time$\downarrow$ & \cellcolor{tspgreen}Obj.$\downarrow$ & \cellcolor{tspgreen}Gap$\downarrow$ & \cellcolor{tspgreen}Time$\downarrow$ \\ \midrule
Concorde & 10.72 & 0.00\% & 3.0m & 16.55 & 0.00\% & 37.6m & 23.12 & 0.00\% & 7.9h & 50.96 & 0.00\% & 18.4h \\
LKH-3 & 10.72 & 0.00\% & 0.8m & 16.55 & 0.00\% & 1.5m & 23.12 & 0.00\% & 24m & 50.96 & 0.00\% & 9.6h \\ \midrule
AM & 11.03 & 2.89\% & 0.3m & 21.24 & 28.34\% & 1.4m & 35.18 & 52.16\% & 2.2m & 96.37 & 89.11\% & 5.7m \\
POMO & 10.97 & 2.33\% & 0.2m & 20.85 & 25.98\% & 1.0m & 32.94 & 42.47\% & 1.6m & 87.79 & 72.27\% & 4.8m \\
LEHD & 10.79 & 0.65\% & 0.3m & 17.08 & 3.20\% & 0.8m & 24.70 & 6.83\% & 2.5m & 55.82 & 9.54\% & 8.6m \\
CNF & 10.80 & 0.74\% & 1.5m & 17.29 & 4.47\% & 3.3m & 25.19 & 8.95\% & 7.7m & 56.53 & 10.93\% & 11.1m \\
GFlowNet-HBG & \underline{10.78} & 0.56\% & 0.1m & \underline{17.05} & 3.02\% & 0.5m & \underline{24.42} & 5.62\% & 1.3m & \underline{54.67} & 7.28\% & 3.7m \\
ECO w/o LS Boot. & 10.96 & 2.23\% & 0.1m & 17.73 & 7.12\% & 0.4m & 25.33 & 9.55\% & 0.9m & 57.02 & 11.89\% & 2.5m \\
\textbf{ECO} & \textbf{10.76} & \textbf{0.37\%} & \textbf{0.1m} & \textbf{16.98} & \textbf{2.59\%} & \textbf{0.4m} & \textbf{24.24} & \textbf{4.84\%} & \textbf{0.9m} & \textbf{53.76} & \textbf{5.49\%} & \textbf{2.5m} \\ \midrule
 & \multicolumn{3}{c|}{\cellcolor{cvrpblue}CVRP100} & \multicolumn{3}{c|}{\cellcolor{cvrpblue}CVRP200} & \multicolumn{3}{c|}{\cellcolor{cvrpblue}CVRP500} & \multicolumn{3}{c}{\cellcolor{cvrpblue}CVRP1000} \\
 & \cellcolor{cvrpblue}Obj.$\downarrow$ & \cellcolor{cvrpblue}Gap$\downarrow$ & \cellcolor{cvrpblue}Time$\downarrow$ & \cellcolor{cvrpblue}Obj.$\downarrow$ & \cellcolor{cvrpblue}Gap$\downarrow$ & \cellcolor{cvrpblue}Time$\downarrow$ & \cellcolor{cvrpblue}Obj.$\downarrow$ & \cellcolor{cvrpblue}Gap$\downarrow$ & \cellcolor{cvrpblue}Time$\downarrow$ & \cellcolor{cvrpblue}Obj.$\downarrow$ & \cellcolor{cvrpblue}Gap$\downarrow$ & \cellcolor{cvrpblue}Time$\downarrow$ \\ \midrule
HGS & 15.56 & 0.00\% & 0.4m & 19.63 & 0.00\% & 1.0m & 37.15 & 0.00\% & 6.1m & 63.26 & 0.00\% & 13.7m \\ \midrule
AM & 16.73 & 7.51\% & 0.2m & 21.33 & 8.64\% & 0.8m & 47.36 & 27.39\% & 2.0m & 104.40 & 65.03\% & 3.9m \\
POMO & 16.15 & 3.79\% & 0.2m & 20.47 & 4.26\% & 0.6m & 46.13 & 24.17\% & 1.4m & 100.19 & 58.37\% & 2.1m \\
LEHD & 15.82 & 1.67\% & 0.3m & 20.11 & 2.44\% & 0.6m & 38.72 & 4.23\% & 2.0m & 67.08 & 6.04\% & 4.5m \\
CNF & 15.85 & 1.86\% & 0.3m & 20.08 & 2.31\% & 1.8m & 38.51 & 3.64\% & 4.3m & 66.94 & 5.82\% & 8.7m \\
GFlowNet-HBG & \underline{15.70} & 0.91\% & 0.1m & \underline{20.01} & 1.96\% & 0.2m & \textbf{38.19} & \textbf{2.81\%} & 0.7m & \underline{65.09} & 2.89\% & 1.7m \\
ECO w/o LS Boot. & 15.87 & 1.99\% & 0.1m & 20.13 & 2.57\% & 0.2m & 38.61 & 3.94\% & 0.6m & 67.11 & 6.08\% & 1.2m \\
\textbf{ECO} & \textbf{15.69} & \textbf{0.86\%} & \textbf{0.1m} & \textbf{19.66} & \textbf{1.77\%} & \textbf{0.2m} & \underline{38.21} & 2.85\% & \textbf{0.6m} & \textbf{65.08} & \textbf{2.87\%} & \textbf{1.8m} \\ \bottomrule
\end{tabular}%
}
\end{table*}

\subsection{Experimental Setup}

\textbf{Instances and Protocol.} We evaluate TSP on $N \in \{200,500,1000,5000\}$ and CVRP on $N \in \{100,200,500,1000\}$, using the same 1,000 held-out test instances per scale for all methods. ECO and all retrained neural baselines are trained and evaluated at the same target scale. Main-table neural results use single-policy inference without test-time local search; method-specific multi-start or sampling budgets are listed in Table~\ref{tab:eval_protocol}. To separate model throughput from batching policy, we report both a max-utilization protocol in Table~\ref{tab:main_results} and a fixed-batch protocol in Table~\ref{tab:fixed_batch_runtime}.

\textbf{Model and Training.} ECO uses embedding dimension 128 with 3-layer Mamba encoder and decoder. We set $\beta=0.3$, $K=32$, update the reference model every 10 iterations, and train with Adam at learning rate $5 \times 10^{-4}$. Training consists of 10 SFT epochs on 100,000 LKH-labeled instances followed by batched DPO with LS-augmented preference pairs ($\alpha=0.3$); LS is used only during training. Neural experiments run on a single NVIDIA A800 GPU (80GB). Traditional solvers run on a 32-thread Intel Xeon CPU node with one solver process per test split.

\textbf{Baselines.} We compare against Concorde, LKH, HGS, and representative neural baselines AM, POMO, LEHD, CNF, and GFlowNet-HBG. AM/POMO/LEHD are retrained at each reported scale when an official pretrained checkpoint at that scale is unavailable; CNF and GFlowNet-HBG use their official implementations and recommended inference budgets. We also discuss BQ, GLOP, DIFUSCO, NeuOpt, and augmentation-based solvers as related scalable NCO families because their inference pipelines include heatmap decoding, partitioning, diffusion refinement, or iterative improvement that is not directly comparable to a pure constructive forward pass. In addition, for controlled comparisons on TSP1000 we construct matched counterfactual variants that share the same embedding size, decoding protocol, data budget, and stopping criterion: \textit{Transformer+DPO}, \textit{Mamba+PPO}, \textit{Mamba+pairwise logistic ranking}, \textit{Mamba+hinge ranking}, and \textit{SFT-only}. For offline objectives, the compared methods use the same buffered candidate pool and the same LS augmentation ratio; for PPO, we use the same rollout budget and the same Mamba backbone unless otherwise stated.
\vspace{-5pt}
\subsection{Experimental Analysis}
\vspace{-5pt}
To systematically address the core objectives of this study, we structure our experimental analysis around the following three Research Questions (RQs):

RQ1 (Scalability and Quality): Can ECO simultaneously achieve strong solution quality and high efficiency on large-scale instances compared with mainstream constructive NCO baselines?

RQ2 (Efficiency Mechanism): Does the proposed Mamba architecture empirically reduce sequence-modeling memory growth and improve throughput over attention mechanisms in long-sequence tasks?

RQ3 (Component Effectiveness): What are the individual contributions of the Mamba backbone, the DPO objective, the warm-up strategy, and heuristic bootstrapping, and how do they compare with alternative pairwise objectives under matched training budgets?

\begin{table}[t]
    \caption{\textbf{Fixed-batch timing audit.} Runtime in minutes on the same 1,000-instance test sets with inference batch size fixed to 8 for all neural methods. The max-utilization time is copied from Table~\ref{tab:main_results}. Objective values are unchanged by the batching protocol.}
    \label{tab:fixed_batch_runtime}
    \vspace{-5pt}
    \centering
    \resizebox{\columnwidth}{!}{
    \begin{tabular}{lcccc}
        \toprule
        Method & \multicolumn{2}{c}{TSP1000 Time} & \multicolumn{2}{c}{CVRP500 Time} \\
        & Max & Fixed-8 & Max & Fixed-8 \\
        \midrule
        AM & 2.2 & 2.8 & 2.0 & 2.4 \\
        POMO & 1.6 & 2.0 & 1.4 & 1.8 \\
        LEHD & 2.5 & 3.1 & 2.0 & 2.3 \\
        CNF & 7.7 & 7.9 & 4.3 & 4.4 \\
        GFlowNet-HBG & 1.3 & 1.5 & 0.7 & 0.8 \\
        \textbf{ECO} & \textbf{0.9} & \textbf{1.0} & \textbf{0.6} & \textbf{0.7} \\
        \bottomrule
    \end{tabular}}
\end{table}

\begin{table}[t]
    \caption{\textbf{Bootstrap 95\% confidence intervals for close neural comparisons.} We report mean gap $\pm$ 95\% CI over 10,000 bootstrap resamples of the 1,000 test instances.}
    \label{tab:ci_close}
    \vspace{-5pt}
    \centering
    \resizebox{\columnwidth}{!}{
    \begin{tabular}{lccc}
        \toprule
        Method & TSP1000 & CVRP500 & CVRP1000 \\
        \midrule
        CNF & $8.95{\pm}0.09$ & $3.64{\pm}0.08$ & $5.82{\pm}0.12$ \\
        GFlowNet-HBG & $5.62{\pm}0.07$ & $\textbf{2.81}{\pm}0.06$ & $2.89{\pm}0.07$ \\
        \textbf{ECO} & $\textbf{4.84}{\pm}0.05$ & $2.85{\pm}0.05$ & $\textbf{2.87}{\pm}0.08$ \\
        \bottomrule
    \end{tabular}}
\end{table}

\subsubsection{Main Results Analysis (Responds to RQ1)}
\begin{figure*}[!ht]  
\centering
\includegraphics[width=\linewidth]{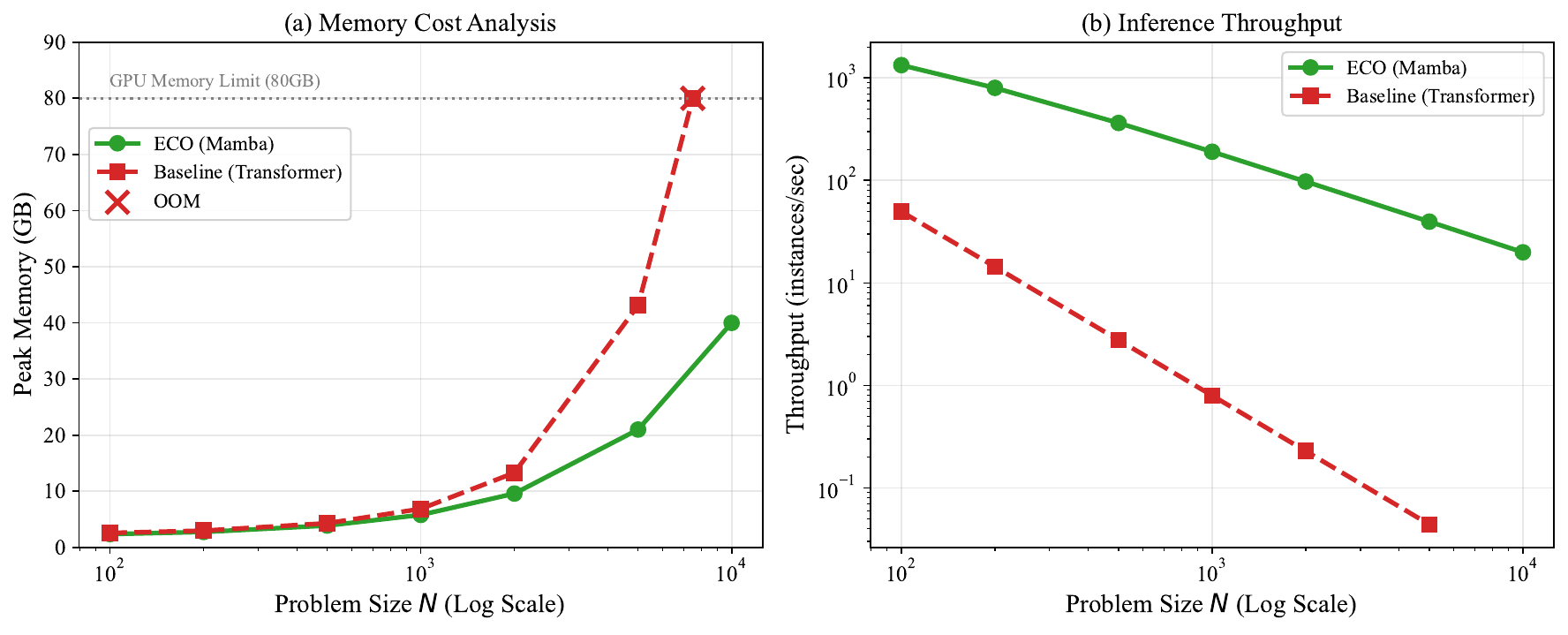}
\vspace{-0.3cm} 

\caption{
\textbf{Memory and throughput comparison on an NVIDIA A800 GPU.}
(a) Peak memory usage versus problem size. ECO scales near-linearly and avoids the Transformer's memory explosion. (b) Autoregressive inference throughput. ECO sustains substantially higher throughput as the sequence length increases.
}
\label{fig:efficiency}
\vspace{-0.4cm}
\end{figure*}
Table~\ref{tab:main_results} presents the comparative performance on TSP and CVRP. The results support four key observations:

\textbf{ECO is Best or Competitive Across Most Settings:} As shown in Table~\ref{tab:main_results}, ECO obtains the best results among all compared neural methods on every reported TSP scale and on most CVRP scales, while remaining competitive on CVRP500 where GFlowNet-HBG is slightly better. The confidence intervals in Table~\ref{tab:ci_close} show that the CVRP500/CVRP1000 differences are small, whereas ECO's TSP1000 advantage over the strongest neural baselines is larger than the interval width. These results support the claim that the proposed training and architecture design improves solution quality without merely trading quality for speed.

\textbf{Large-Scale Robustness Against Attention Collapse:} The most critical validation of our approach is observed on the largest instances (TSP5000/CVRP1000). As shown in Table~\ref{tab:main_results}, standard Transformer-based baselines (AM, POMO) suffer severe degradation as the problem size increases, with gaps reaching $89.11\%$ and $72.27\%$ on TSP5000. In contrast, ECO maintains a much smaller gap of $5.49\%$ on TSP5000 while preserving a short evaluation time. This indicates that the proposed architecture and training pipeline remain much more stable when scaling to thousands of nodes.

\textbf{Training-Time LS Bootstrapping Yields a Better Pure Policy:} The gap between `ECO' and `ECO w/o LS Boot.' directly measures the effect of LS-guided preference construction, because neither row uses local search during inference. The full ECO model consistently outperforms the ablated version on both TSP and CVRP, indicating that LS provides a stronger training signal that is effectively distilled into the policy parameters rather than acting as a test-time repair heuristic.

\textbf{Inference Latency and Practicality:} ECO demonstrates a clear efficiency advantage among the compared constructive neural solvers. On TSP5000, ECO completes the 1,000-instance evaluation in 2.5 minutes, which is faster than CNF and GFlowNet-HBG while requiring much less wall-clock time than traditional exact or heavy heuristic solvers. This establishes ECO as a practical option for time-sensitive large-scale routing scenarios. Notably, this speed is achieved without any local-search post-processing at test time.

\subsubsection{Permutation Sensitivity Analysis}
\label{sec:perm_sensitivity}

\begin{table}[t]
    \caption{\textbf{Sensitivity to input node order.} For each trained model, we evaluate 16 random permutations of the same test instances and report mean gap and the standard deviation across permutations. POMO is included as a set-equivariant attention reference.}
    \label{tab:perm_sensitivity}
    \vspace{-5pt}
    \centering
    \resizebox{\columnwidth}{!}{
    \begin{tabular}{lcccc}
        \toprule
        Method / order & \multicolumn{2}{c}{TSP1000 Gap} & \multicolumn{2}{c}{CVRP500 Gap} \\
        & Mean & Perm. std. & Mean & Perm. std. \\
        \midrule
        POMO & 42.47 & 0.02 & 24.17 & 0.03 \\
        ECO, raw input order & 5.31 & 0.43 & 3.17 & 0.28 \\
        ECO, $x$-coordinate sort & 4.97 & 0.18 & 3.01 & 0.14 \\
        ECO, Hilbert order & \textbf{4.82} & 0.04 & \textbf{2.83} & 0.04 \\
        ECO, Morton order (ours) & 4.84 & \textbf{0.01} & 2.85 & \textbf{0.01} \\
        \bottomrule
    \end{tabular}}
\end{table}

Because Mamba is not permutation-equivariant by construction, we explicitly evaluate order sensitivity instead of assuming set invariance. Table~\ref{tab:perm_sensitivity} shows that using the raw sampling order introduces visible variance under random re-indexing. Spatial canonicalization substantially reduces this effect. Hilbert and Morton orderings obtain similar gaps, while Morton gives negligible permutation variance because random input permutations are mapped back to the same canonical sequence before encoding. We use Morton order in the main model because it provides nearly the same quality as Hilbert ordering with simpler and faster preprocessing.

\subsubsection{Efficiency and Throughput Analysis (Responds to RQ2)}
\label{sec:4.2.2}


To isolate the efficiency gains attributed to the proposed SSM architecture, we conducted a controlled stress test. We implemented a control-variate baseline by replacing the Mamba backbone within the ECO framework with a standard Transformer architecture while keeping the rest of the constructive pipeline identical. Importantly, all throughput numbers correspond to true autoregressive inference rather than teacher-forced routed decoding.

\textbf{GPU Memory Scaling Analysis:} We monitored the peak GPU memory consumption across problem sizes $N \in \{100, \dots, 10000\}$. To isolate architectural memory growth, especially encoder activations and recurrent decoder state, we standardized the inference batch size to 8 for all measurements.

\textbf{Results:} As illustrated in Figure \ref{fig:efficiency} (Left), the Transformer baseline exhibits a quadratic memory increase from full attention, triggering an Out-Of-Memory (OOM) error on the A800 (80GB) GPU at approximately $N=7000$. Conversely, ECO maintains a near-linear memory growth profile for sequence modeling and recurrent decoding state. Even at $N=10,000$, the memory footprint remains within hardware limits (approx. 40GB). This confirms that the Mamba encoder-decoder substantially reduces the memory bottleneck preventing Transformer-based models from scaling, while the action scorer still performs dense candidate comparisons as discussed in Sec.~\ref{sec:complexity}.

\textbf{Limit-State Throughput Testing:} To evaluate the maximum processing capacity suitable for industrial deployment, we adopted a Maximal Utilization Protocol for throughput testing. Instead of a fixed batch size, we dynamically maximized the inference batch size for each model and problem size until the GPU memory limit was reached.

\textbf{Results:} Figure~\ref{fig:efficiency} (Right) reports throughput (instances/second) on a logarithmic scale. While the Transformer's throughput plummets for large $N$ because the batch size must be reduced to accommodate the attention matrices, ECO sustains much higher throughput. This gap highlights that ECO's lower memory footprint not only saves VRAM but also permits larger effective batch sizes during inference, which is the main source of the practical speedup.

\subsection{Ablation and Controlled Comparisons (Responds to RQ3)}

\begin{table}[t]
    \caption{\textbf{Objective-level comparison on TSP1000 under a matched Mamba backbone.} All variants use the same SFT initialization, the same preference data budget, the same LS augmentation ratio, and the same stopping criterion. ``Rel.'' indicates whether an explicit reference policy is used during preference optimization.}
    \label{tab:objective_controls}
    \vspace{-5pt}
    \centering
    \resizebox{\columnwidth}{!}{
    \begin{tabular}{lccc}
        \toprule
        Objective & Rel. & Gap (\%) $\downarrow$ & Time $\downarrow$ \\
        \midrule
        Pairwise logistic ranking & No & 5.31 & 10.0h \\
        Margin hinge ranking & No & 5.58 & 9.8h \\
        Online PPO & -- & 4.88 & 34.5h \\
        \textbf{DPO (ours)} & Yes & \textbf{4.84} & \textbf{10.2h} \\
        \bottomrule
    \end{tabular}}
    
\end{table}

\begin{table}[t]
    \caption{\textbf{Factorized ablation and counterfactual controls on TSP1000.} We separately toggle the backbone, iterative update rule, SFT warm-up, and LS bootstrapping. ``--'' indicates divergence or unstable training.}
    \label{tab:factorized_ablation}
    \vspace{-5pt}
    \centering
    \resizebox{\columnwidth}{!}{
    \begin{tabular}{lcccc}
        \toprule
        Variant & Backbone / Update & SFT & LS & Gap (\%) $\downarrow$ / Time $\downarrow$ \\
        \midrule
        SFT-only & Mamba / NLL & \cmark & \xmark & 18.76 / 1.1h \\
        w/o SFT & Mamba / DPO & \xmark & \cmark & $>20.0$ / -- \\
        w/o LS Boot. & Mamba / DPO & \cmark & \xmark & 9.55 / 9.5h \\
        Transformer+DPO & Transformer / DPO & \cmark & \cmark & 5.47 / 14.6h \\
        Mamba+PPO & Mamba / PPO & \cmark & \cmark & 4.88 / 34.5h \\
        Transformer+PPO & Transformer / PPO & \cmark & \xmark & 9.73 / 38.7h \\
        \textbf{ECO (Full)} & \textbf{Mamba / DPO} & \cmark & \cmark & \textbf{4.84 / 10.2h} \\
        \bottomrule
    \end{tabular}}
    \vspace{-18pt}
\end{table}

\begin{table}[t]
    \caption{\textbf{Sensitivity to the LS-augmented pair ratio $\alpha$.} Gap is reported under pure neural inference. Margin denotes the average normalized winner--loser cost margin in the DPO buffer.}
    \label{tab:alpha_sensitivity}
    \vspace{-5pt}
    \centering
    \resizebox{\columnwidth}{!}{
    \begin{tabular}{cccc}
        \toprule
        $\alpha$ & TSP1000 Gap & CVRP500 Gap & Margin (\%) \\
        \midrule
        0.0 & 9.55 & 3.94 & 0.22 \\
        0.1 & 6.21 & 3.24 & 1.05 \\
        0.3 & \textbf{4.84} & \textbf{2.85} & 1.87 \\
        0.5 & 4.92 & 2.88 & 2.19 \\
        1.0 & 5.37 & 3.06 & 3.04 \\
        \bottomrule
    \end{tabular}}
\end{table}

\begin{table}[t]
    \caption{\textbf{Disentangling LS supervision on TSP1000 and CVRP500.} For CVRP, all moves are accepted only when capacity feasibility is preserved; infeasible relocate, swap, or 2-opt* moves are rejected before pair construction.}
    \label{tab:ls_operator_ablation}
    \vspace{-5pt}
    \centering
    \resizebox{\columnwidth}{!}{
    \begin{tabular}{lcc}
        \toprule
        Training signal & TSP1000 Gap & CVRP500 Gap \\
        \midrule
        Raw self-play DPO, no LS & 9.55 & 3.94 \\
        SFT on LS-refined routes only & 7.42 & 3.37 \\
        DPO + 2-opt / relocate & 5.21 & 3.05 \\
        DPO + 3-opt / swap & 5.08 & 2.98 \\
        \textbf{DPO + combined LS (ours)} & \textbf{4.84} & \textbf{2.85} \\
        \bottomrule
    \end{tabular}}
\end{table}

\begin{table}[t]
    \caption{\textbf{Winner--loser margin distribution after LS refinement.} Margins are normalized by the loser cost and reported in percent for the DPO buffer at $\alpha=0.3$.}
    \label{tab:ls_margin_distribution}
    \vspace{-5pt}
    \centering
    \resizebox{\columnwidth}{!}{
    \begin{tabular}{lcccc}
        \toprule
        Pair source & Mean & Median & P10 & P90 \\
        \midrule
        TSP1000 raw self-play & 0.22 & 0.17 & 0.03 & 0.48 \\
        TSP1000 LS-refined & 1.87 & 1.42 & 0.31 & 3.76 \\
        CVRP500 raw self-play & 0.18 & 0.14 & 0.02 & 0.41 \\
        CVRP500 LS-refined & 1.46 & 1.13 & 0.24 & 2.91 \\
        \bottomrule
    \end{tabular}}
\end{table}

To verify the contribution of the training objective, the backbone architecture, the SFT warm-up, and the LS-guided preference construction, we perform two additional controlled studies on TSP1000 beyond the coarse ablation in the main comparison table. The first study fixes the Mamba backbone and all data-generation settings, then varies only the optimization objective. The second study explicitly disentangles four factors: backbone (Transformer vs. Mamba), update rule (no iterative preference update vs. PPO vs. DPO), SFT initialization, and LS bootstrapping.

\textbf{DPO is not interchangeable with generic pairwise losses.} Table~\ref{tab:objective_controls} shows that, under the same Mamba backbone and the same buffered preference data, DPO outperforms both reference-free pairwise logistic ranking and hinge ranking. All three offline objectives are much faster than PPO, but DPO achieves the best final gap. This indicates that the gain is not merely due to converting costs into pairwise labels; rather, the reference-anchored DPO objective provides a more stable continuation path across iterative refresh cycles. In practice, we observe that the reference-free objectives become more sensitive to late-stage label noise when the winner--loser margin narrows, whereas DPO better preserves monotonic improvement.

\textbf{The Mamba backbone contributes beyond the training objective alone.} Table~\ref{tab:factorized_ablation} allows a matched comparison across backbones. Under the same online PPO paradigm, replacing the Transformer with Mamba improves the TSP1000 gap from $9.73\%$ to $4.88\%$ while slightly reducing training time, showing that the linear-memory backbone is already beneficial before introducing DPO. Under the same DPO-style iterative preference update, Transformer+DPO still underperforms the full Mamba-based ECO and takes longer to train, which suggests that the architecture and the batched preference update are complementary rather than redundant.

\textbf{The iterative preference stage is necessary, but only after a competent warm-up.} The comparison between SFT-only and the full ECO model shows that behavior cloning alone provides a usable policy but leaves a large gap ($18.76\%$ vs. $4.84\%$). Adding iterative DPO closes most of this remaining gap. In contrast, removing SFT causes the training to become unstable, confirming that direct preference optimization needs a non-trivial initial policy to produce meaningful winner--loser pairs. This controlled result is consistent with the cold-start explanation in Sec.~3.2.

\textbf{LS bootstrapping supplies a stronger signal than self-play alone.} Comparing ``w/o LS Boot.'' against the full ECO model isolates the effect of stronger preference construction. Without LS, the model still benefits from batched DPO, but its gap deteriorates from $4.84\%$ to $9.55\%$. Table~\ref{tab:alpha_sensitivity} further shows that the benefit is not monotonic in the amount of LS supervision: $\alpha=0.3$ gives the best trade-off, while $\alpha=1.0$ over-concentrates the buffer on locally refined demonstrations and reduces diversity. Table~\ref{tab:ls_operator_ablation} shows that SFT on LS-refined routes is weaker than DPO pairs, supporting the view that ECO benefits from preference margins rather than simply cloning the LS output distribution. Table~\ref{tab:ls_margin_distribution} confirms that LS-refined pairs widen the average winner--loser margin by roughly one order of magnitude over raw self-play pairs. For CVRP, the LS operators include intra-route 2-opt, capacity-preserving relocate/swap moves, and inter-route 2-opt* exchanges; every candidate move is checked against the remaining-capacity constraint before acceptance. Importantly, all rows use the same pure neural inference protocol, so the gain cannot be attributed to any search procedure at deployment time.

\textbf{DPO matches PPO quality at a much lower optimization cost.} With the same Mamba backbone, the gap difference between PPO and DPO is marginal ($4.88\%$ vs. $4.84\%$), but the total training time drops from $34.5$h to $10.2$h. This gap-time trade-off is central to our efficiency claim: the improvement is not only that ECO is fast in inference, but that it also attains PPO-level policy quality with an offline-in-update training loop that is much better aligned with GPU execution.

\textbf{Overall summary.} Taken together, Tables~\ref{tab:objective_controls} and~\ref{tab:factorized_ablation} provide a substantially more fine-grained attribution of the final gain than a single four-row ablation. The results indicate that the full ECO advantage is jointly produced by four ingredients: a scalable Mamba backbone, a reference-anchored DPO objective that is stronger than generic pairwise ranking, an SFT warm-up that resolves the cold-start regime, and LS-guided bootstrapping that widens preference margins during iterative refinement.

\section{Conclusion}

In this paper, we introduced ECO, an efficiency-oriented NCO framework that combines batched preference optimization with a Mamba backbone. By unifying a scalable Mamba encoder-decoder, a two-stage preference-learning pipeline, and LS-guided bootstrapping, ECO achieves best-or-competitive performance among the compared neural baselines while maintaining a markedly better memory/throughput profile on large TSP and CVRP instances. Our complexity analysis also clarifies the boundary of the claim: Mamba reduces encoder memory scaling and recurrent decoding-state cost, whereas naive constructive action scoring remains quadratic. Our results show that LS-guided preference construction improves the learned policy even though local search is used only during training and never during inference. These findings suggest that hardware-aware architectures and batched preference-based optimization provide a strong direction for scalable neural routing solvers.



\appendices

\section{Theoretical Analysis of ECO-DPO}
This appendix strengthens the connection between ECO's practical training loop and its underlying optimization target. We distinguish three levels: the finite candidate pool actually used to build preference pairs, the exact KL-regularized soft policy-improvement target, and the finite-sample policy produced by neural DPO optimization. The results are intentionally stated as approximate guarantees because finite rollouts, non-convex function approximation, and periodic reference refreshes do not support a global monotonicity theorem without additional assumptions.

\textbf{Scope of the theory.} The propositions below are best interpreted as calibration and error-decomposition results, not as a convergence proof for the implemented neural algorithm. Proposition~1 assumes a Bradley--Terry preference model and an expressive distribution over a finite candidate pool. In the actual ECO implementation, labels are produced by deterministic cost comparison. Under fully deterministic and separable labels, an unconstrained logistic objective may drive margins to infinity rather than select a unique finite Gibbs distribution. In practice, finite model capacity, reference anchoring, early stopping, stochastic optimization, and finite candidate pools act as implicit regularizers. Thus the Gibbs result explains the regularized/noisy preference regime to which DPO is calibrated, while the deterministic ECO labels should be viewed as a low-noise ranking limit rather than an exact realization of the proposition.

\subsection{Candidate-Set DPO Consistency}
For an instance $X$, let $\mathcal{Y}(X)$ be the feasible solution set and let $\mathcal{C}(y|X)$ be the cost to be minimized. At one DPO refresh, ECO samples a finite candidate pool $\mathcal{S}=\{y_1,\ldots,y_K\}\subset\mathcal{Y}(X)$. Let $q_0(y|X)$ denote the reference policy restricted and renormalized on $\mathcal{S}$, with $q_0(y|X)>0$ for all $y\in\mathcal{S}$. For $\lambda>0$, define the restricted Gibbs policy
\begin{equation}
    q_\lambda(y|X,\mathcal{S})
    =
    \frac{q_0(y|X)\exp[-\lambda\mathcal{C}(y|X)]}
    {\sum_{z\in\mathcal{S}}q_0(z|X)\exp[-\lambda\mathcal{C}(z|X)]}.
\label{eq:restricted_gibbs}
\end{equation}

\textbf{Proposition 1 (finite-pool DPO consistency).}
Assume preferences inside $\mathcal{S}$ follow the Bradley--Terry model
\begin{equation}
    P(y_a\succ_X y_b)
    =
    \sigma\left(
    \frac{\mathcal{C}(y_b|X)-\mathcal{C}(y_a|X)}{\tau}
    \right),
\end{equation}
where $\tau>0$, and the pair sampler assigns positive probability to every unordered pair in $\mathcal{S}$. Consider the population DPO risk over distributions $q\in\Delta(\mathcal{S})$ with score
\begin{equation}
    h_q(y)=\beta\log\frac{q(y|X)}{q_0(y|X)} .
\end{equation}
If the model class can represent any distribution on $\mathcal{S}$, the Bayes-optimal DPO distribution is the restricted Gibbs policy in Eq.~\eqref{eq:restricted_gibbs} with $\lambda=1/(\beta\tau)$. Consequently,
\begin{equation}
    \mathcal{C}(y_a|X)<\mathcal{C}(y_b|X)
    \Longrightarrow
    \frac{q_\lambda(y_a|X,\mathcal{S})}{q_0(y_a|X)}
    >
    \frac{q_\lambda(y_b|X,\mathcal{S})}{q_0(y_b|X)} .
\end{equation}

\emph{Proof.}
For a fixed pair $(y_a,y_b)$, the Bernoulli logistic risk is minimized when the predicted logit equals the true log-odds:
\begin{equation}
    h_q(y_a)-h_q(y_b)
    =
    \operatorname{logit}P(y_a\succ_X y_b)
    =
    \frac{\mathcal{C}(y_b|X)-\mathcal{C}(y_a|X)}{\tau}.
\end{equation}
This system of pairwise equations is consistent because it is realized by $h^*(y)=-\mathcal{C}(y|X)/\tau+c(X,\mathcal{S})$. Since $h_q(y)=\beta\log(q(y|X)/q_0(y|X))$, normalization over $\mathcal{S}$ gives
\begin{equation}
    q^*(y|X,\mathcal{S})
    \propto
    q_0(y|X)\exp[-\mathcal{C}(y|X)/(\beta\tau)],
\end{equation}
which is Eq.~\eqref{eq:restricted_gibbs} with $\lambda=1/(\beta\tau)$. The ranking statement follows immediately from the monotonicity of the exponential tilt. \hfill $\square$

This result bridges the all-solution Gibbs view and ECO's real finite candidate pools: population DPO on a sampled pool is calibrated to the Gibbs improvement restricted to that pool when preferences are noisy and Bradley--Terry calibrated. The deterministic winner--loser labels used in ECO preserve the same cost-induced ordering but do not, by themselves, identify a unique finite Gibbs optimum without regularization or capacity constraints.

\subsection{Approximate Soft Policy-Improvement Bound}
Let $\pi_t$ be the reference policy at outer iteration $t$. The full soft-improvement target for an instance is
\begin{equation}
    \pi_{t,\lambda}^*(y|X)
    =
    \frac{\pi_t(y|X)\exp[-\lambda\mathcal{C}(y|X)]}
    {Z_{t,\lambda}(X)} .
\end{equation}
For the random candidate-pool sampler used by ECO, let $q_{t,\lambda}^{\mathcal{S}}$ be the restricted target in Eq.~\eqref{eq:restricted_gibbs}, and define the population candidate-pool target
\begin{equation}
    \bar{q}_{t,K,\lambda}(\cdot|X)
    =
    \mathbb{E}_{\mathcal{S}}\left[
    q_{t,\lambda}^{\mathcal{S}}(\cdot|X)
    \right],
\end{equation}
where each restricted distribution is treated as a distribution on $\mathcal{Y}(X)$ with zero mass outside $\mathcal{S}$. Let
\begin{equation}
\begin{split}
    \epsilon_t
    &=
    \mathbb{E}_{X}
    D_{\mathrm{TV}}\big(
    \pi_{t+1}(\cdot|X),\bar{q}_{t,K,\lambda}(\cdot|X)
    \big),\\
    \delta_t
    &=
    \mathbb{E}_{X}
    D_{\mathrm{TV}}\big(
    \bar{q}_{t,K,\lambda}(\cdot|X),\pi_{t,\lambda}^*(\cdot|X)
    \big).
\end{split}
\end{equation}
Here $\epsilon_t$ summarizes optimization, estimation, and function-approximation error in DPO, while $\delta_t$ summarizes the bias caused by using finite candidate pools instead of the full feasible set.

\textbf{Proposition 2 (approximate soft improvement).}
Assume the per-instance cost range is bounded by $R$, i.e.,
\begin{equation}
    \sup_{X}\left[
    \max_{y}\mathcal{C}(y|X)-\min_{y}\mathcal{C}(y|X)
    \right]\le R .
\end{equation}
Let $J(\pi)=\mathbb{E}_{X,y\sim\pi(\cdot|X)}\mathcal{C}(y|X)$. Then
\begin{equation}
    J(\pi_{t+1})
    \le
    J(\pi_t)
    -
    I_t(\lambda)
    +
    R(\epsilon_t+\delta_t),
\label{eq:approx_improvement_bound}
\end{equation}
where
\begin{equation}
    I_t(\lambda)
    =
    \mathbb{E}_{X}
    \int_0^\lambda
    \operatorname{Var}_{y\sim\pi_{t,s}^*(\cdot|X)}
    \left[\mathcal{C}(y|X)\right] ds
    \ge 0 .
\end{equation}
Therefore, the round improves the expected cost whenever
$I_t(\lambda)>R(\epsilon_t+\delta_t)$.

\emph{Proof.}
For a fixed $X$, differentiating
$\bar{\mathcal{C}}_s(X)=
\mathbb{E}_{y\sim\pi_{t,s}^*}\mathcal{C}(y|X)$ gives
\begin{equation}
    \frac{d}{ds}\bar{\mathcal{C}}_s(X)
    =
    -
    \operatorname{Var}_{y\sim\pi_{t,s}^*}
    \left[\mathcal{C}(y|X)\right].
\end{equation}
Integrating from $0$ to $\lambda$ yields
$J(\pi_{t,\lambda}^*)=J(\pi_t)-I_t(\lambda)$. For any two distributions $p,q$ over the same feasible set and any cost range bounded by $R$,
\begin{equation}
    \left|
    \mathbb{E}_{p}\mathcal{C}
    -
    \mathbb{E}_{q}\mathcal{C}
    \right|
    \le
    R D_{\mathrm{TV}}(p,q).
\end{equation}
Applying this inequality and the triangle decomposition through $\bar{q}_{t,K,\lambda}$ gives Eq.~\eqref{eq:approx_improvement_bound}. \hfill $\square$

Proposition 2 makes the role of each ECO component explicit. The soft target contributes the ideal improvement $I_t(\lambda)$. Larger or better-covered candidate pools reduce $\delta_t$. More accurate DPO optimization and a more expressive policy class reduce $\epsilon_t$. We do not estimate $\epsilon_t$ or $\delta_t$ exactly in the experiments because they involve distances to inaccessible population targets over the full feasible set. The theorem therefore should not be read as an empirical monotonic-improvement certificate. It states the condition under which the approximate finite-pool update would inherit the cost-decreasing property of the exact soft target and provides a diagnostic interpretation of candidate-pool coverage and optimization error.

\subsection{Gradient Structure of the DPO Surrogate}
For a sampled pair $(y_w,y_l)$, define
\begin{equation}
\begin{split}
    s_\theta(X,y_w,y_l)=
    \beta\Big[
    &\log\frac{\pi_\theta(y_w|X)}{\pi_t(y_w|X)}
    -
    \log\frac{\pi_\theta(y_l|X)}{\pi_t(y_l|X)}
    \Big].
\end{split}
\end{equation}
The single-pair DPO gradient is
\begin{equation}
\begin{split}
    \nabla_\theta \ell_{\mathrm{DPO}}
    =
    -\beta\sigma(-s_\theta)
    \Big[
    &\nabla_\theta\log\pi_\theta(y_w|X)\\
    &-
    \nabla_\theta\log\pi_\theta(y_l|X)
    \Big].
\end{split}
\end{equation}
Thus DPO implements a reference-relative likelihood-ratio update that pushes probability mass from higher-cost candidates to lower-cost candidates. Because $\log\pi_\theta(y|X)$ decomposes autoregressively over construction steps, this pairwise signal is distributed across the route decisions that produced the two solutions.
\vspace{-5pt}
\subsection{Local-Search Margin and Statistical Signal}
Let $y_{\mathrm{raw}}\sim\pi_t(\cdot|X)$ and let
$y_{\mathrm{ref}}=\mathcal{T}_{\mathrm{LS}}(y_{\mathrm{raw}})$ be the locally refined solution used only during training. Define the LS improvement margin
\begin{equation}
    m_{\mathrm{LS}}(X)
    =
    \mathcal{C}(y_{\mathrm{raw}}|X)
    -
    \mathcal{C}(y_{\mathrm{ref}}|X)
    \ge 0 .
\end{equation}

\textbf{Proposition 3 (margin improves preference signal).}
Under the Bradley--Terry model with temperature $\tau$, let
$p_m=P(y_{\mathrm{ref}}\succ_X y_{\mathrm{raw}})=\sigma(m_{\mathrm{LS}}/\tau)$.
Then:
\begin{equation}
    P(\text{wrong label})
    =
    1-p_m
    =
    \sigma(-m_{\mathrm{LS}}/\tau)
    \le
    \exp(-m_{\mathrm{LS}}/\tau),
\end{equation}
\begin{equation}
\begin{split}
    &\left|
    \frac{d}{ds}
    \left[
    -p_m\log\sigma(s)
    -(1-p_m)\log\sigma(-s)
    \right]_{s=0}
    \right| \\
    &\qquad =
    \frac{1}{2}\tanh\left(\frac{m_{\mathrm{LS}}}{2\tau}\right),
\end{split}
\end{equation}
and, for the signed preference variable $Z\in\{+1,-1\}$ with
$P(Z=+1)=p_m$,
\begin{equation}
    \frac{|\mathbb{E}Z|}{\sqrt{\operatorname{Var}(Z)}}
    =
    \sinh\left(\frac{m_{\mathrm{LS}}}{2\tau}\right).
\end{equation}
All three quantities improve monotonically as $m_{\mathrm{LS}}$ increases.

\emph{Proof.}
The error bound follows from $\sigma(-a)\le e^{-a}$ for $a\ge0$. The derivative identity follows by differentiating the binary logistic risk at neutral score $s=0$. Finally, $2p_m-1=\tanh(m_{\mathrm{LS}}/(2\tau))$ and
$\operatorname{Var}(Z)=1-(2p_m-1)^2=\operatorname{sech}^2(m_{\mathrm{LS}}/(2\tau))$, which gives the stated signal-to-noise ratio. \hfill $\square$

Proposition 3 formalizes the benefit of LS bootstrapping beyond the statement that the refined route has lower cost. A larger local-search margin exponentially suppresses preference-label ambiguity, increases the initial logistic gradient magnitude for an uncalibrated pair, and improves the directional signal-to-noise ratio. Hence LS-augmented pairs reduce the effective statistical difficulty of pairwise learning while still leaving inference purely neural.
\vspace{-5pt}
\section{Detailed Implementation Settings}

To ensure the reproducibility of our results, we provide the detailed hyperparameter settings for both the neural architecture and the training phases.

\subsection{Network Architecture}
Our Mamba-based Encoder-Decoder follows the official implementation configurations of Mamba~\cite{mamba}. The specific architectural hyperparameters are detailed in Table \ref{tab:arch_params}.

\begin{table}[h]
\caption{Hyperparameters of the Mixed Mamba Architecture.}
\vspace{-5pt}
\label{tab:arch_params}
\centering  
\begin{sc}
\begin{tabularx}{\columnwidth}{Xc} 
\toprule
Parameter & Value \\
\midrule
Embedding Dimension ($D$) & 128 \\
Number of Layers ($L$) & 3 \\
Mamba State Dimension ($d_{state}$) & 16 \\
Mamba Conv Kernel Size ($d_{conv}$) & 4 \\
Mamba Expansion Factor ($E$) & 2 \\
Feed-Forward Network (FFN) Ratio & 4 \\
Normalization & RMSNorm \\
\bottomrule
\vspace{-10pt}
\end{tabularx}
\end{sc}
\end{table}

\vspace{-15pt}
\subsection{Baseline Configurations and Reproducibility}

To ensure a fair comparison, all neural baselines were evaluated on the same NVIDIA A800 GPU and the same 1,000 test instances per scale. Table~\ref{tab:eval_protocol} summarizes the decoding and batching settings used for the main results. We utilized the official open-source implementations for the learning-based methods when available:

\begin{table}[h]
\caption{\textbf{Evaluation protocol for neural baselines.} ``Max batch'' means the largest batch fitting in 80GB GPU memory at the target scale. Fixed-batch timing uses batch size 8 for all listed methods.}
\vspace{-5pt}
\label{tab:eval_protocol}
\centering
\resizebox{\columnwidth}{!}{
\begin{tabular}{llll}
\toprule
Method & Decoding / budget & Batch policy & Model source \\
\midrule
AM & greedy, 1 rollout & fixed-8 / max batch & retrained per scale \\
POMO & native multi-start, no 8-fold aug. & fixed-8 / max batch & retrained per scale \\
LEHD & heavy-decoder greedy & fixed-8 / max batch & official or retrained \\
CNF & official flow steps & fixed-8 / max batch & official recipe \\
GFlowNet-HBG & 32 sampled candidates & fixed-8 / max batch & official 24h budget \\
ECO & greedy, 1 rollout, no LS & fixed-8 / max batch & retrained per scale \\
\bottomrule
\end{tabular}}
\end{table}

\begin{itemize}
    \item \textbf{Attention Model (AM)} and \textbf{POMO}: We used the implementation from the widely adopted \texttt{RL4CO} library\footnote{\url{https://github.com/ai4co/rl4co}}. We retrained them at the corresponding scale when no official checkpoint was available and used validation-tuned batch sizes within the same 80GB GPU limit.
    \item \textbf{LEHD}: We used the official LEHD implementation~\cite{5}; for scales without released checkpoints, we retrained using the authors' recommended heavy-decoder setting.
    \item \textbf{GFlowNet-HBG}: We utilized the official code provided by \cite{GFlowNet}, training the model with their recommended settings for 24 hours and using the default 32-candidate inference budget.
    \item \textbf{CNF}: We reproduced the results using the official code from \cite{CNF}, ensuring the flow-matching steps were consistent with their paper.
\end{itemize}

For traditional solvers: 
\begin{itemize}
    \item \textbf{Concorde}: We used the PyConcorde wrapper on CPU.
    \item \textbf{LKH-3}: We used the official binary with standard parameters (runs=1, max-trials=10000).
    \item \textbf{HGS}: We used the public HGS-CVRP implementation with its default stopping criterion for each instance size.
\end{itemize}
All traditional solvers were run on the same 32-thread Intel Xeon CPU node. We did not mix CPU and GPU time in the neural runtime columns; solver times are reported as wall-clock CPU time.
\vspace{-10pt}
\subsection{Data Generation and Local Search Details}
Following standard NCO protocols \cite{3}, problem instances for both TSP and CVRP are generated by sampling $N$ node coordinates uniformly from the unit square $[0, 1]^2$. For CVRP, demands are sampled uniformly from discrete values $\{1, \dots, 9\}$. The capacities used in the main experiments are $Q=50,80,150,250$ for $N=100,200,500,1000$, respectively. For compatibility with smaller-scale prior protocols, we use $Q=30$ and $Q=40$ for $N=20$ and $N=50$ when those scales are used for pretraining or debugging.

In the bootstrapping DPO phase, we employ local search operators to refine raw model outputs \emph{only for training-time preference construction}. No local search is used during inference. For TSP, we use first-improvement 2-opt followed by bounded 3-opt. For CVRP, we use intra-route 2-opt, customer relocate, customer swap, and inter-route 2-opt*; every move is accepted only if all route capacities remain feasible and the total route length does not increase. To balance efficiency and solution quality during data generation:
\begin{itemize}
\item We implement a First Improvement strategy rather than Best Improvement to reduce computational overhead.
\item The maximum number of accepted local-search moves is capped at $N$ for TSP and $2N$ for CVRP to prevent excessive runtime during the offline data generation process.
\item This operator is implemented using Numba JIT compilation to ensure it does not become a bottleneck in the training pipeline.
\item During DPO training, we incorporate preference pairs enhanced by local search to strengthen the learning signal. Specifically, these augmented pairs constitute 30\% of the total training data in the main setting.
\end{itemize}

\vspace{-5pt}
\bibliographystyle{IEEEtran}
\bibliography{ref}

\end{document}